\documentclass{article}
\usepackage{spconf,amsmath,graphicx}
\usepackage{multirow}
\usepackage[table,xcdraw]{xcolor}
\usepackage{amsmath}
\usepackage{mathtools}

\newcommand{\argmax}[1]{\underset{#1}{\operatorname{arg}\,\operatorname{max}}\;}
\newcommand{\argmin}[1]{\underset{#1}{\operatorname{arg}\,\operatorname{min}}\;}
\title{Partial Domain Adaptation Using Graph Convolutional Networks}
\name{Seunghan Yang \qquad Youngeun Kim \qquad Dongki Jung \qquad Changick Kim}
\address{Korea Advanced Institute of Science and Technology (KAIST), Daejeon, Korea}
\begin{document}
%
\maketitle
\begin{abstract}
Partial domain adaptation (PDA), in which we assume the target label space is included in the source label space, is a general version of standard domain adaptation. Since the target label space is unknown, the main challenge of PDA is to reduce the learning impact of irrelevant source samples, named outliers, which do not belong to the target label space. Although existing partial domain adaptation methods effectively down-weigh outliers' importance, they do not consider data structure of each domain and do not directly align the feature distributions of the same class in the source and target domains, which may lead to misalignment of category-level distributions. To overcome these problems, we propose a {\it graph partial domain adaptation} (GPDA) network, which exploits Graph Convolutional Networks for jointly considering data structure and the feature distribution of each class. Specifically, we propose a {\it label relational graph} to align the distributions of the same category in two domains and introduce {\it moving average centroid separation} for learning networks from the {\it label relational graph}. We demonstrate that considering data structure and the distribution of each category is effective for PDA and our GPDA network achieves state-of-the-art performance on the Digit and Office-31 datasets.

\end{abstract}

\section{Introduction}
\label{sec:intro}

Recently, deep learning-based methods have shown state-of-the-art performance in image classification beyond human perception. However, these methods require a lot of labeled data to train deep neural networks. Since it costs a lot of time and money to obtain labels for training, there is a limitation to applying them to real situations.
Unsupervised domain adaptation methods \cite{ganin2014unsupervised, kim2019delegated, das2018unsupervised, fu2019improved} have received attention as a way to reduce the labeling cost. They aim to ensure a network trained with rich labeled data from the source domain working well on unlabeled data from the target domain. As the source data and the target data are sampled from different distributions, the networks trained on the source domain without domain adaptation do not infer well on the target samples. To bridge different domains, most domain adaptation methods try to learn domain invariant feature representations by adversarial learning \cite{ganin2014unsupervised}. These methods successfully reduce the large gap between different domains, and domain adaptation methods for image classification \cite{kim2019delegated, das2018unsupervised, fu2019improved, xu2019larger, cao2018partial, saito2019semi} perform well as much as the network trained on rich labeled data from the target domain.

\begin{figure}

\begin{minipage}[b]{1.0\linewidth}
  \centering
  \centerline{\includegraphics[width=4.5cm]{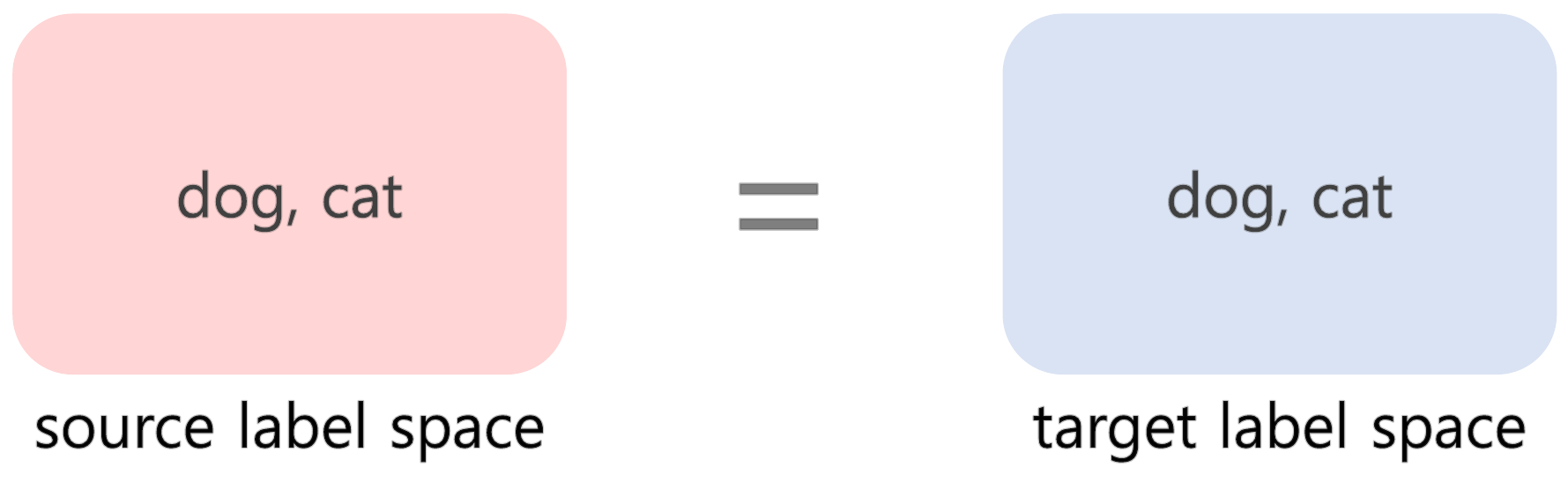}}
  \centerline{\small(a) Standard domain adaptation}\medskip
\end{minipage}
\begin{minipage}[b]{1.0\linewidth}
  \centering
  \centerline{\includegraphics[width=4.5cm]{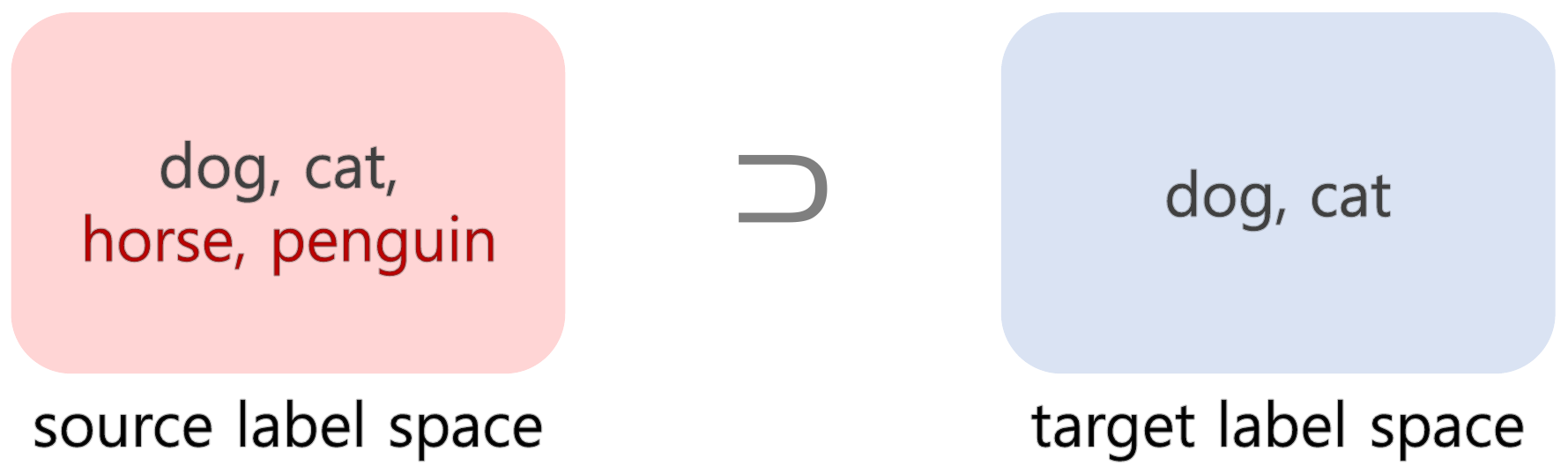}}
  \centerline{\small(b) Partial domain adaptation}\medskip
\end{minipage}
\vspace{-8mm}
\caption{Example of comparison with standard domain adaptation and partial domain adaptation.} 
\label{fig:res}
\vspace{-4mm}
\end{figure}

\begin{figure*}[htb]
\begin{minipage}[b]{1.0\linewidth}
  \centering
  \centerline{\includegraphics[width=13cm]{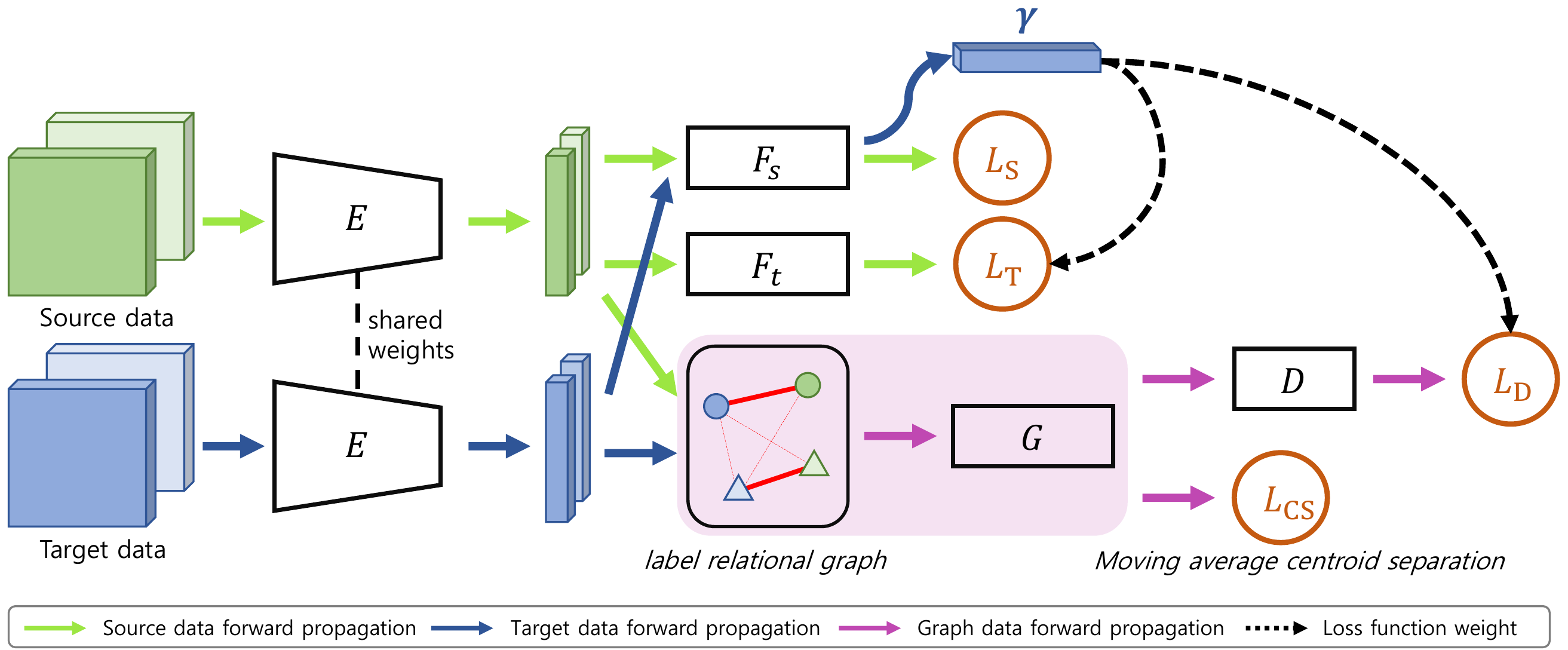}}
\end{minipage}
\caption{Our architecture, named the GPDA network, aims to reduce the impact of the outliers by considering data structure. We form a {\it label relational graph} using source and target data, and graph data is propagated with GCN to align the distributions of the same class. We introduce {\it moving average centroid separation} to give constraint, which maps graph features in the different classes separately.}
\label{fig:res}
\vspace{-5mm}
\end{figure*}

However, existing domain adaptation methods generally assume that the source and target domains share the identical label space. Under this assumption, when the source and target datasets have different label spaces, the domain difference can not be correctly reduced. In real applications, it is not practical to find or generate a source domain with the identical label space as the target domain. To overcome this problem, partial domain adaptation (PDA) has been studied under the assumption that the target label space is included in the source label space. In PDA, we use a large dataset with numerous classes as a source dataset and transfer source domain knowledge to a small target domain with few categories. PDA is feasible for many applications because large datasets with many classes are open to the public so that we can use it as a source domain, and classes of the target dataset are highly likely to be subsets of the source dataset.

Partial domain adaptation is more challenging than standard domain adaptation since the target label space is unknown, and there are irrelevant source samples, named outliers, which do not belong to the target label space. Therefore, most partial domain adaptation methods try to prevent learning with outliers. Cao {\it et al.} propose the PADA \cite{cao2018partial} and SAN \cite{cao2018partial1} architectures to down-weigh outliers' importance automatically by introducing the probabilities of source samples belonging to the target label space. Similarly, Zhang {\it et al.} suggest IWAN \cite{zhang2018importance} by exploiting an additional domain classifier and Cao {\it et al.} design ETN \cite{cao2019learning}, which adds a new classifier and a novel domain classifier to select outliers accurately. Matsuura {\it et al.} propose TWINs \cite{matsuura2018twins}, which estimate the ratio of the target samples in each class for weighting the classes present in the target domain. Although these novel methods \cite{cao2018partial, cao2018partial1, zhang2018importance, cao2019learning, matsuura2018twins} effectively perform partial domain adaptation by reducing the effect of outliers on training, they do not consider data structure of each domain, which is known to reflect the marginal or conditional distribution \cite{long2014transfer} and data statistic ~\cite{zhang2017joint, xu2015discriminative}, and they do not directly align the feature distributions of the same category in the source and target domains.

In this paper, we construct a graph for data structure to align the feature distributions of the same category for partial domain adaptation. 
Specifically, we propose a {\it label relational graph} exploiting the relationship between pseudo labels for the target samples and ground truth labels for the source samples. Moreover, we introduce {\it moving average centroid separation} for learning networks from the {\it label relational graph}. By using the {\it label relational graph} and {\it moving average centroid separation}, the features of the same classes in the source and target domains are incorporated together, while the features of the different classes are separated from each other. To consider these two modules jointly, we propose a {\it graph partial domain adaptation} (GPDA) network.
Our network is effective for partial domain adaptation and that achieves state-of-the-art performance on the Digit \cite{lecun1998gradient, hull1994database} and Office-31 \cite{saenko2010adapting} datasets.


\section{Proposed Method}
\label{sec:majhead}
In partial domain adaptation, we are given by $n_{s}$ source samples and the corresponding class labels $\left \{(x_{i}^{s}, y_{i}^{s})\right \}_{i=1}^{n_{s}}$ from the source domain $D_{s}$ with $\left | C_{s} \right |$ classes and $n_{t}$ unlabeled target samples $\left \{x_{i}^{t}\right \}_{i=1}^{n_{t}}$ from the target domain $D_{t}$ with $\left | C_{t} \right |$ classes, where $\left | C_{s} \right |$ denotes the source label space whereas $\left | C_{t} \right |$ the target label space. Like standard domain adaptation, we assume that the source and target samples are drawn from different probability distribution $p$ and $q$, respectively. Moreover, in partial domain adaptation, $C_{t}$ is the subset of $C_{s}$, {\it i.e.}, $C_{t} \subset C_{s}$, and $C_{t}$ is unknown. In other words, we know the target label space is included in the source label space, but do not know what kinds of classes are included in the target label space.
\subsection{Graph Partial Domain Adaptation}
\label{sssec:subhead}
In partial domain adaptation, adopting standard domain adaptation algorithms, which learn all source and target samples with the same weight, cause performance degeneration by outliers, {\it i.e.}, $x_{i}^{s}$ corresponding to $y_{i}^{s} \notin \left | C_{t} \right |$. Therefore, we introduce a {\it graph partial domain adaptation} (GPDA) network, which aims to reduce the learning impact of the outliers during training time. As illustrated in Fig. 2, we exploit data structure, forcing to align the distributions of the same category in a weight framework. $E$ is a feature extractor, and $F_{s}$ and $F_{t}$ are classifiers for source samples in the source label space and the common label space, respectively. Task-specific features are learned by classifiers in supervised learning as followed:
\begin{align}
L_{S}(\theta _{{F}_{s}}, \theta _{E})=\frac{1}{n_{s}}\sum_{x_{i}\in D_{s}}L_{y}(F_{s}(E(x_{i})), y_{i}), \\
L_{T}(\theta _{{F}_{t}}, \theta _{E})=\frac{1}{n_{s}}\sum_{x_{i}\in D_{s}}\gamma _{{y_{i}}}L_{y}(F_{t}(E(x_{i})), y_{i}),
\end{align}
where $L_{y}$ is the cross entropy loss function, and $\gamma_{{y_{i}}}$ is a probability of a source label $y_{i}$ belonging to the target label space. Here, $\boldsymbol{\gamma} \in R^{\left | C_{s} \right |}$, which indicates the contribution of each source class, and $\boldsymbol{\gamma}$ can be calculated as follows:
\begin{equation}
\boldsymbol{\gamma} = \frac{1}{n_{t}}\sum_{x_{i}\in D_{t}}F_{s}(E(x_{i})).
\end{equation}
For domain invariant features, we combine Graph Convolutional Networks $G$ with a domain classifier $D$. $D$ is trained to distinguish the source domain from the target domain, and simultaneously $G$ and $E$ are trained to confuse $D$. This loss function can be expressed as follows:
\begin{equation}
\begin{split}
L_{D}(\theta _{E}, \theta _{G}, \theta _{D}) = & -\frac{1}{n_{s}}\sum_{x_{i}\in D_{s}}\gamma _{{y_{i}}}L_{bce}(D(G(E(x_{i}), A), d_{i})) \\
& -\frac{1}{n_{t}}\sum_{x_{i}\in D_{t}}L_{bce}(D(G(E(x_{i}), A), d_{i})), 
\end{split}
\end{equation}
where $L_{bce}$ is the binary cross entropy loss and $d_{i}\in \{0,1\}$ indicates the domain label.
Moreover, a {\it label relational graph} $A$ and {\it moving average centroid separation} $L_{CS}$ lead to aligning the distributions of same category in the source and target domains, which are described in section 2.2 and section 2.3, respectively. The total objective function is as follows:
\begin{equation}
\begin{aligned}
L_{GPDA} = L_{S} + L_{T}+ \lambda _{1}L_{D}+ \lambda _{2}L_{CS},
\end{aligned}
\end{equation}
where $\lambda _{1}$ and $\lambda _{2}$ are the trade-off parameters, and we set trade-off parameters to $\lambda_{1} = 1.0$ and $\lambda_{2} = 1.0$ for all experiments. Finally, the proposed GPDA network can be solved by a minimax optimization problem as follows:
\begin{align}
\begin{aligned}
(\hat \theta _{E}, \hat \theta _{G}, \hat \theta _{{F}_{s}}, \hat \theta _{{F}_{t}}) & = \argmin{\theta _{E}, \theta _{G}, \theta _{{F}_{s}}, \theta _{{F}_{t}}}L_{GPDA}, \\
\hat \theta _{D} & = \argmax{\theta _{D}}L_{GPDA}.
\end{aligned}
\end{align}

\subsection{GCN with A Label Relational Graph}
\label{ssec:subhead}
Graph Convolutional Networks (GCN)\cite{kipf2016semi} are motivated by a first-order approximation of localized spectral filters on graphs ~\cite{hammond2011wavelets}. 
Each GCN layer with $N$ nodes is described as followed:
\begin{equation}
Z = {\tilde{D}}^{-\frac{1}{2}}\tilde{A}{\tilde{D}}^{-\frac{1}{2}}X\Theta,
\end{equation}
where $X\in {R}^{N\times F}$ is a $F$-dimensional node signal matrix, and $\Theta \in {R}^{F\times F'}$ is a learnable filter changing a node signal to $F'$ dimension with $\tilde{A} = A + I_{N}$ and $\tilde{D}_{ii} = \sum_{j}\tilde{A}_{ij}$.

GCN outperforms in tasks on datasets with defined node-to-node relationships \cite{sen2008collective}, and has recently been used in computer vision. Unlike datasets such as Citeseer, Cora, and Pubmed \cite{sen2008collective}, which are datasets with predefined nodes and edges, it is important to determine nodes and edges appropriately for tasks in computer vision. To adopt Graph Convolutional Networks for computer vision tasks, Chen {\it et al.} \cite{chen2019multi} use the graph by setting nodes to word embedding vectors and edges to class co-occurrence patterns within the dataset for multi-label image classification. In addition, for group action recognition, Wu {\it et al.} \cite{wu2019learning} set nodes to feature maps of people and edges to appearance relation. However, since there is no graph for partial domain adaptation, we construct a {\it label relational graph} to use Graph Convolutional Networks in partial domain adaptation. 

In the {\it label relational graph}, each node feature represents the feature map of a sample, then the {\it i}th node feature of the graph $X_{i}$ is obtained by:
\vspace{-1mm}
\begin{equation}
X_i = E(x_{i}),
\end{equation}
\vspace{-1mm}
where $E$ and $x_{i}$ indicate the feature extractor and the {\it i}th input image, respectively. An adjacency matrix $A$ contains relationships of the nodes, {\it i.e.}, edges, and each component of $A$ is obtained as follows:
\begin{equation}
{A}_{ij} = \sum_{c=0}^{C-1}{y_{i,c}y_{j,c}},
\end{equation}
where $C$ is the number of class, and $y_{i,c}$ and $y_{j,c}$ are labels of the {\it i}th and {\it j}th image, respectively. In the case of the source images, there are corresponding ground truth labels, which are one-hot vectors, but in the case of the target images, labels are not given in training. We exploit pseudo-labels \cite{lee2013pseudo}, which are well known for semi-supervised learning techniques. 
Finally, the {\it label relational graph} has a large value between images that are likely to have the same class and has a low value for unrelated images. Layerwise propagation of the GCN layer with the {\it label relational graph} smooths features of images with the same class, which leads to aligning the distributions of the same class.

\begin{table}[]
\caption{The classification accuracy of the Digit dataset in the partial domain adaptation setting.}
\centering
\resizebox{6.1cm}{!}{
\begin{tabular}{ccc}
\hline
\multicolumn{1}{c}{}                                  & \multicolumn{2}{c}{Digit}                                                                                                                    \\ \cline{2-3} 
\multicolumn{1}{c}{\multirow{-2}{*}{Method}} & MNIST $\rightarrow$ USPS & USPS $\rightarrow$ MNIST \\ \hline
Source only & 85.2 & 80.0 \\
DANN~\cite{ganin2014unsupervised} & 67.1 & 72.1 \\
IWAN \cite{zhang2018importance} & 90.6 & 85.7 \\
TWINs \cite{matsuura2018twins}& 96.3 & 90.2 \\ \hline
GPDA  & \textbf{96.9} & \textbf{94.6} \\ \hline
\end{tabular}
}
\vspace{-4mm}
\end{table}

\begin{table*}[]
\caption{The classification accuracy of the Office-31 dataset in the partial domain adaptation setting.}
\centering
\resizebox{10.5cm}{!}{
\begin{tabular}{cccccccc}
\hline
\multicolumn{1}{c}{}                                  & \multicolumn{7}{c}{Office-31}                                                                                                                    \\ \cline{2-8} 
\multicolumn{1}{c}{\multirow{-2}{*}{Method}} & A $\rightarrow$ W & D $\rightarrow$ W            & W $\rightarrow$ D & A $\rightarrow$ D & D $\rightarrow$ A & W $\rightarrow$ A & Average        \\ \hline
ResNet \cite{he2016deep}                                                 & 75.59             & {\color[HTML]{000000} 96.27} & 98.09             & 83.44             & 83.92             & 84.97             & 87.05          \\
DANN \cite{ganin2014unsupervised}                                                   & 73.56             & 96.27                        & 98.73             & 81.53             & 82.78             & 86.12             & 86.50          \\
IWAN \cite{zhang2018importance}                                                   & 89.15             & 99.32                        & 99.36             & 90.45             & 95.62             & 94.26             & 94.69          \\
SAN \cite{cao2018partial1}                                                    & 93.90             & 99.32                        & 99.36             & 94.27             & 94.15             & 88.73             & 94.96          \\
PADA \cite{cao2018partial}                                                   & 86.54             & 99.32                        & \textbf{100}      & 82.17             & 92.69             & 95.41             & 92.69          \\
ETN \cite{cao2019learning}                                                   & 94.52             & \textbf{100}                 & \textbf{100}      & 95.03             & \textbf{96.21}    & 94.64             & 96.73          \\ \hline
Baseline      & 88.81    & \textbf{100}                 & \textbf{100}      & 94.27    & 88.73             & 94.89 & 94.45 \\
Ours w/o $L_{CS}$   & 94.58    & \textbf{100}                 & \textbf{100}      & 92.36    & 94.26             & 94.89    & 96.02 \\
Ours w/o graph                                                    & 95.59    & \textbf{100}                 & \textbf{100}      & 94.27    & 94.26             & 94.89    & 96.50 \\
Ours (GPDA)                                                    & \textbf{96.95}    & \textbf{100}                 & \textbf{100}      & \textbf{98.73}    & 95.10             & \textbf{95.83}    & \textbf{97.77} \\ \hline
\end{tabular}
}
\vspace{-2mm}
\end{table*}

\subsection{Moving Average Centroid Separation}
\label{sssec:subhead} 
Graph convolution may lead to smoothing the features of different classes because of incorrect pseudo labels. To alleviate this smoothing effect, we introduce {\it moving average centroid separation}, which follows the idea in \cite{xie2018learning}. Specifically, we use features of the labeled source samples and pseudo-labeled target samples. In \cite{xie2018learning}, they design the class centroid alignment module to map features in the same class nearby, while we introduce {\it moving average centroid separation} to map features in the different classes separately. The {\it moving average centroid separation} objective function is as follows:
\begin{equation}
L_{CS}(\theta _{E}, \theta _{G}) = -\sum_{k=0}^{C-1}\parallel c^{s}_{k} - c^{t}_{(k+i)\,mod\,C} \parallel ^{2},
\end{equation}
where $c^{s}_{k}$ and $c^{t}_{k}$ are centroids of feature maps of class $k$ in the source and target domains, respectively, and $i$ is a random integer number within [1, $C-1$], updated in each iteration. Through the objective function, false signals in pseudo-labeled target samples are suppressed, and features in the different classes are explicitly separated from each other.

\section{Experiments}
\label{sec:print}
\vspace{-1mm}
\subsection{Setup}
\vspace{-1mm}
We evaluate our architecture to compare with state-of-the-art networks for partial domain adaptation on two benchmark datasets: Digit~\cite{lecun1998gradient, hull1994database} and Office-31~\cite{saenko2010adapting}. 

{\bf Digit.}
We utilize MNIST and USPS for two domain adaptation tasks ({\it i.e.}, {\bf MNIST} $\rightarrow$ {\bf USPS} and {\bf USPS} $\rightarrow$ {\bf MNIST}).
MNIST and USPS consist of 10 images containing numbers from 0 to 9, but domains are different. MNIST is collected from students, whereas USPS is a dataset for US portal service. In the PDA setting, we use all images and labels as the source dataset, and adopt images corresponding to the first five classes as the target dataset as conducted in \cite{matsuura2018twins} ({\it i.e.}, $\left | C_{s} \right |=10$, $\left | C_{t} \right | = 5$).

{\bf Office-31.}
Office-31 is a standard benchmark for domain adaptation. It consists of 4,652 images and 31 categories collected from three different domains. {\it Amazon} ({\bf A}) contains images from amazon.com. {\it DSLR} ({\bf D}) and {\it Webcam} ({\bf W}) are taken by a DSLR camera and a web camera, respectively. We utilize the Office-31 dataset for six domain adaptation tasks. 
We use all images and labels as the source dataset, and adopt images corresponding to the ten classes as the target dataset as conducted in~\cite{cao2018partial1} ({\it i.e.}, $\left | C_{s} \right |=31$, $\left | C_{t} \right | = 10$).

We implement our GPDA network based on Pytorch, and exploit CNN architectures for the Digit dataset, as the same protocol in DANN~\cite{ganin2014unsupervised}. For the office-31 dataset, we finetune ResNet-50~\cite{he2016deep} pre-trained on ImageNet~\cite{russakovsky2015imagenet}. For a fair comparison, we use the same base network for previous methods. We add two GCN layers with 256 and 1024 channels on Digit and Office-31, respectively, since increasing the number of layers or channels for GCN layers do not improve performance. New layers are trained from scratch with 10 times the learning rate of the pre-trained layer. We use SGD with the momentum of 0.9 and the learning rate decay strategy implemented in DANN~\cite{ganin2014unsupervised}, and the learning rates of all new layers are increased gradually from 0 to 1 as DANN~\cite{ganin2014unsupervised} also.

\vspace{-3mm}
\subsection{Results and Analysis}

We show that our GPDA network outperforms previous methods on the Digit dataset in Table 1. Source only and DANN are the methods without the domain adaptation algorithm and with the standard domain adaptation algorithm, respectively. In the situation where the label spaces of the source and target are different, the performance of DANN is lower than the way without using the domain adaptation method, {\it i.e., source only}. This is the result of the outliers interfering with distribution alignment. Our method achieves state-of-the-art performance compared to previous partial domain adaptation methods \cite{zhang2018importance, matsuura2018twins}. 

In Table 2, the proposed method achieves state-of-the-art performance with average gain of 1\% beyond ETN \cite{cao2019learning}, which adds the novel domain classifier. Specifically, our network has the same algorithm for reducing the learning impact of outliers as PADA \cite{cao2018partial}, but shows a performance improvement of 4.6\% on average, exploiting the newly introduced a {\it label relational graph} and {\it moving average centroid separation}. It shows that using data structure and considering the distribution of each category are valid for partial domain adaptation. Moreover, we experiment with ablation studies to examine the effect of each module. Baseline indicates the network without a {\it label relational graph} and {\it moving average centroid separation}. It outperforms PADA by exploiting the additional classifier expected to learn images in the common label space. We experiment our GPDA network without a {\it label relational graph} and {\it moving average centroid separation}, respectively. Ours w/o $L_{CS}$ only leads to align the distributions of the same category, whereas ours w/o graph separate the distributions of different categories. They provide higher performances than baseline, meaning that each module is effective for PDA, and we can see that two modules work complement each other to obtain better overall results.

\section{Conclusion}
\label{sec:copyright}
In this paper, we design a novel architecture, named a {\it graph partial domain adaptation} (GPDA) network, to consider data structure and the distribution of each class for partial domain adaptation. Specifically, we integrate Graph Convolutional Networks into a down-weight framework, and propose a {\it label relational graph} and {\it moving average centroid separation} for graph learning. The experimental results show that our GPDA network outperforms previous state-of-the-art methods, demonstrating the effectiveness of our method.

\clearpage

\bibliographystyle{ieee}
\bibliography{refs}

\begin{thebibliography}{10}

\bibitem{ganin2014unsupervised}
Yaroslav Ganin and Victor Lempitsky,
\newblock ``Unsupervised domain adaptation by backpropagation,''
\newblock {\em arXiv preprint arXiv:1409.7495}, 2014.

\bibitem{kim2019delegated}
Dongwan Kim, Seungmin Lee, Namil Kim, and Seong-Gyun Jeong,
\newblock ``Delegated adversarial training for unsupervised domain
  adaptation,''
\newblock in {\em ICIP}, 2019.

\bibitem{das2018unsupervised}
Debasmit Das and CS~George Lee,
\newblock ``Unsupervised domain adaptation using regularized hyper-graph
  matching,''
\newblock in {\em ICIP}, 2018.

\bibitem{fu2019improved}
Jiahui Fu, Xiaofu Wu, Suofei Zhang, and Jun Yan,
\newblock ``Improved open set domain adaptation with backpropagation,''
\newblock in {\em ICIP}, 2019.

\bibitem{xu2019larger}
Ruijia Xu, Guanbin Li, Jihan Yang, and Liang Lin,
\newblock ``Larger norm more transferable: An adaptive feature norm approach
  for unsupervised domain adaptation,''
\newblock in {\em ICCV}, 2019.

\bibitem{cao2018partial}
Zhangjie Cao, Lijia Ma, Mingsheng Long, and Jianmin Wang,
\newblock ``Partial adversarial domain adaptation,''
\newblock in {\em ECCV}, 2018.

\bibitem{saito2019semi}
Kuniaki Saito, Donghyun Kim, Stan Sclaroff, Trevor Darrell, and Kate Saenko,
\newblock ``Semi-supervised domain adaptation via minimax entropy,''
\newblock {\em arXiv preprint arXiv:1904.06487}, 2019.

\bibitem{cao2018partial1}
Zhangjie Cao, Mingsheng Long, Jianmin Wang, and Michael~I Jordan,
\newblock ``Partial transfer learning with selective adversarial networks,''
\newblock in {\em CVPR}, 2018.

\bibitem{zhang2018importance}
Jing Zhang, Zewei Ding, Wanqing Li, and Philip Ogunbona,
\newblock ``Importance weighted adversarial nets for partial domain
  adaptation,''
\newblock in {\em CVPR}, 2018.

\bibitem{cao2019learning}
Zhangjie Cao, Kaichao You, Mingsheng Long, Jianmin Wang, and Qiang Yang,
\newblock ``Learning to transfer examples for partial domain adaptation,''
\newblock in {\em CVPR}, 2019.

\bibitem{matsuura2018twins}
Toshihiko Matsuura, Kuniaki Saito, and Tatsuya Harada,
\newblock ``Twins: Two weighted inconsistency-reduced networks for partial
  domain adaptation,''
\newblock {\em arXiv preprint arXiv:1812.07405}, 2018.

\bibitem{long2014transfer}
Mingsheng Long, Jianmin Wang, Guiguang Ding, Jiaguang Sun, and Philip~S Yu,
\newblock ``Transfer joint matching for unsupervised domain adaptation,''
\newblock in {\em CVPR}, 2014.

\bibitem{zhang2017joint}
Jing Zhang, Wanqing Li, and Philip Ogunbona,
\newblock ``Joint geometrical and statistical alignment for visual domain
  adaptation,''
\newblock in {\em CVPR}, 2017.

\bibitem{xu2015discriminative}
Yong Xu, Xiaozhao Fang, Jian Wu, Xuelong Li, and David Zhang,
\newblock ``Discriminative transfer subspace learning via low-rank and sparse
  representation,''
\newblock {\em IEEE Transactions on Image Processing}, vol. 25, no. 2, pp.
  850--863, 2015.

\bibitem{lecun1998gradient}
Yann LeCun, L{\'e}on Bottou, Yoshua Bengio, Patrick Haffner, et~al.,
\newblock ``Gradient-based learning applied to document recognition,''
\newblock {\em Proceedings of the IEEE}, vol. 86, no. 11, pp. 2278--2324, 1998.

\bibitem{hull1994database}
Jonathan~J. Hull,
\newblock ``A database for handwritten text recognition research,''
\newblock {\em IEEE Transactions on pattern analysis and machine intelligence},
  vol. 16, no. 5, pp. 550--554, 1994.

\bibitem{saenko2010adapting}
Kate Saenko, Brian Kulis, Mario Fritz, and Trevor Darrell,
\newblock ``Adapting visual category models to new domains,''
\newblock in {\em ECCV}. Springer, 2010.

\bibitem{kipf2016semi}
Thomas~N Kipf and Max Welling,
\newblock ``Semi-supervised classification with graph convolutional networks,''
\newblock {\em arXiv preprint arXiv:1609.02907}, 2016.

\bibitem{hammond2011wavelets}
David~K Hammond, Pierre Vandergheynst, and R{\'e}mi Gribonval,
\newblock ``Wavelets on graphs via spectral graph theory,''
\newblock {\em Applied and Computational Harmonic Analysis}, vol. 30, no. 2,
  pp. 129--150, 2011.

\bibitem{sen2008collective}
Prithviraj Sen, Galileo Namata, Mustafa Bilgic, Lise Getoor, Brian Galligher,
  and Tina Eliassi-Rad,
\newblock ``Collective classification in network data,''
\newblock {\em AI magazine}, vol. 29, no. 3, pp. 93--93, 2008.

\bibitem{chen2019multi}
Zhao-Min Chen, Xiu-Shen Wei, Peng Wang, and Yanwen Guo,
\newblock ``Multi-label image recognition with graph convolutional networks,''
\newblock in {\em CVPR}, 2019.

\bibitem{wu2019learning}
Jianchao Wu, Limin Wang, Li~Wang, Jie Guo, and Gangshan Wu,
\newblock ``Learning actor relation graphs for group activity recognition,''
\newblock in {\em CVPR}, 2019.

\bibitem{lee2013pseudo}
Dong-Hyun Lee,
\newblock ``Pseudo-label: The simple and efficient semi-supervised learning
  method for deep neural networks,''
\newblock in {\em Workshop on Challenges in Representation Learning, ICML}.

\bibitem{he2016deep}
Kaiming He, Xiangyu Zhang, Shaoqing Ren, and Jian Sun,
\newblock ``Deep residual learning for image recognition,''
\newblock in {\em CVPR}, 2016.

\bibitem{xie2018learning}
Shaoan Xie, Zibin Zheng, Liang Chen, and Chuan Chen,
\newblock ``Learning semantic representations for unsupervised domain
  adaptation,''
\newblock in {\em ICML}, 2018.

\bibitem{russakovsky2015imagenet}
Olga Russakovsky et~al.,
\newblock ``Imagenet large scale visual recognition challenge,''
\newblock {\em International journal of computer vision}, vol. 115, no. 3, pp.
  211--252, 2015.

\end{thebibliography}

\end{document}